\documentclass[10pt,twocolumn,letterpaper]{article}

\usepackage{cvpr}
\usepackage{times}
\usepackage{epsfig}
\usepackage{graphicx}
\usepackage{amsmath}
\usepackage{amssymb}

 \normalsize

\usepackage[pagebackref=true,breaklinks=true,letterpaper=true,colorlinks,bookmarks=false]{hyperref}

 \cvprfinalcopy 

\def\cvprPaperID{1625} 

\ifcvprfinal\fi
\begin{document}

\title{Competitive Multi-scale Convolution}

\author{Zhibin Liao \qquad Gustavo Carneiro\\
ARC Centre of Excellence for Robotic Vision\thanks{This research was supported by the Australian Research Council Centre of Excellence for Robotic Vision (project number CE140100016)}\\
University of Adelaide, Australia\\
{\tt\small \{zhibin.liao,gustavo.carneiro\}@adelaide.edu.au}
}

\maketitle

\begin{abstract}

In this paper, we introduce a new deep convolutional neural network (ConvNet) module that promotes competition among a set of multi-scale convolutional filters.  This new module is inspired by the inception module, where we replace the original collaborative pooling stage (consisting of a concatenation of the multi-scale filter outputs) by a competitive pooling represented by a maxout activation unit.  This extension has the following two objectives: 1) the selection of the maximum response among the multi-scale filters prevents filter co-adaptation and allows the formation of multiple sub-networks within the same model, which has been shown to facilitate the training of complex learning problems; and 2) the maxout unit reduces the dimensionality of the outputs from the multi-scale filters.  We show that the use of our proposed module in typical deep ConvNets produces classification results that are either better than or comparable to the state of the art on the following benchmark datasets: MNIST, CIFAR-10, CIFAR-100 and SVHN.

\end{abstract}

\section{Introduction}

The use of competitive activation units in deep convolutional neural networks (ConvNets) is generally understood as a way of building one network by the combination of multiple sub-networks, where each one is capable of solving a simpler task when compared to the complexity of the original problem involving the whole dataset~\cite{srivastava2014understanding}.  Similar ideas have been explored in the past using multi-layer perceptron models~\cite{jacobs1991adaptive}, but there is a resurgence in the use of competitive activation units in deep ConvNets~\cite{srivastava2013compete,srivastava2014understanding}.
For instance, rectified linear unit (ReLU)~\cite{glorot2011deep} promotes a competition between the input sum (usually computed from the output of convolutional layers) and a fixed value of 0, while maxout~\cite{goodfellow2013maxout} and local winner-take-all (LWTA)~\cite{srivastava2013compete} explore an explicit competition among the input units.  As shown by Srivastava et al.~\cite{srivastava2014understanding}, these competitive activation units allow the formation of sub-networks that respond similarly to similar input patterns, which facilitates training~\cite{glorot2011deep,goodfellow2013maxout,srivastava2013compete} and generally produces superior classification results~\cite{srivastava2014understanding}.

\begin{figure}
\begin{center}
\begin{tabular}{c}
\includegraphics[width=0.4\textwidth]{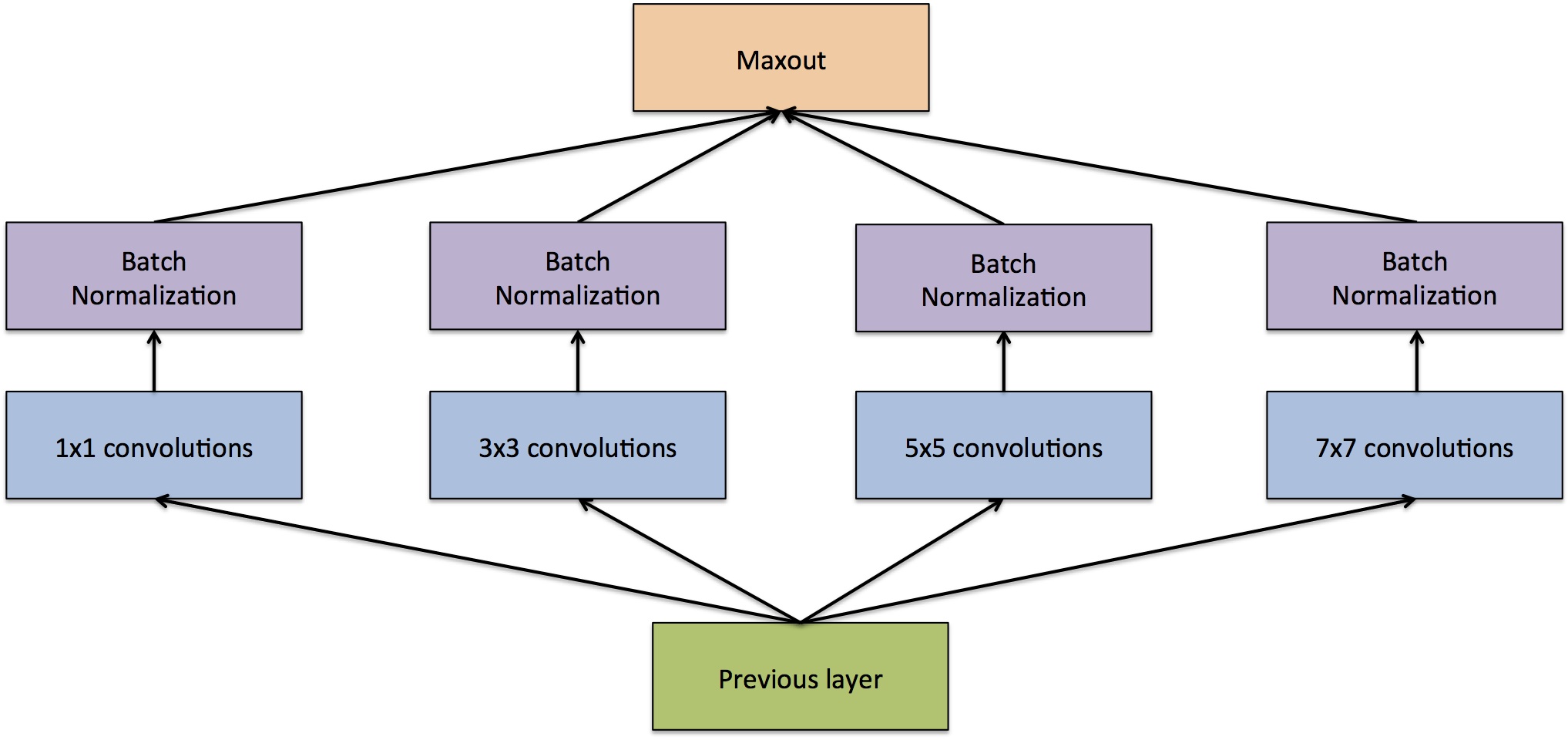} \\
(a) Competitive multi-scale convolution module \\
\includegraphics[width=0.4\textwidth]{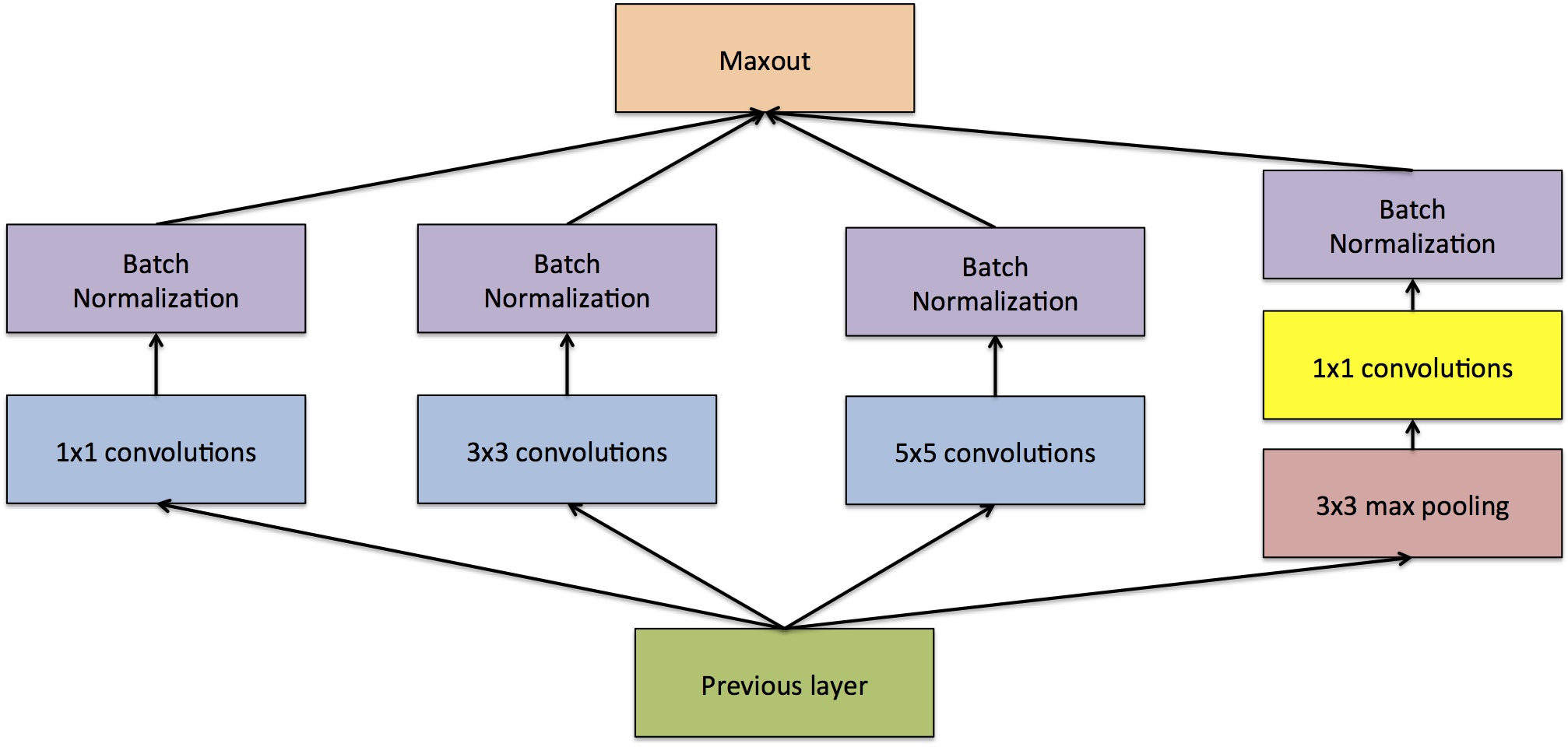} \\
(b) Competitive Inception module\\
\includegraphics[width=0.4\textwidth]{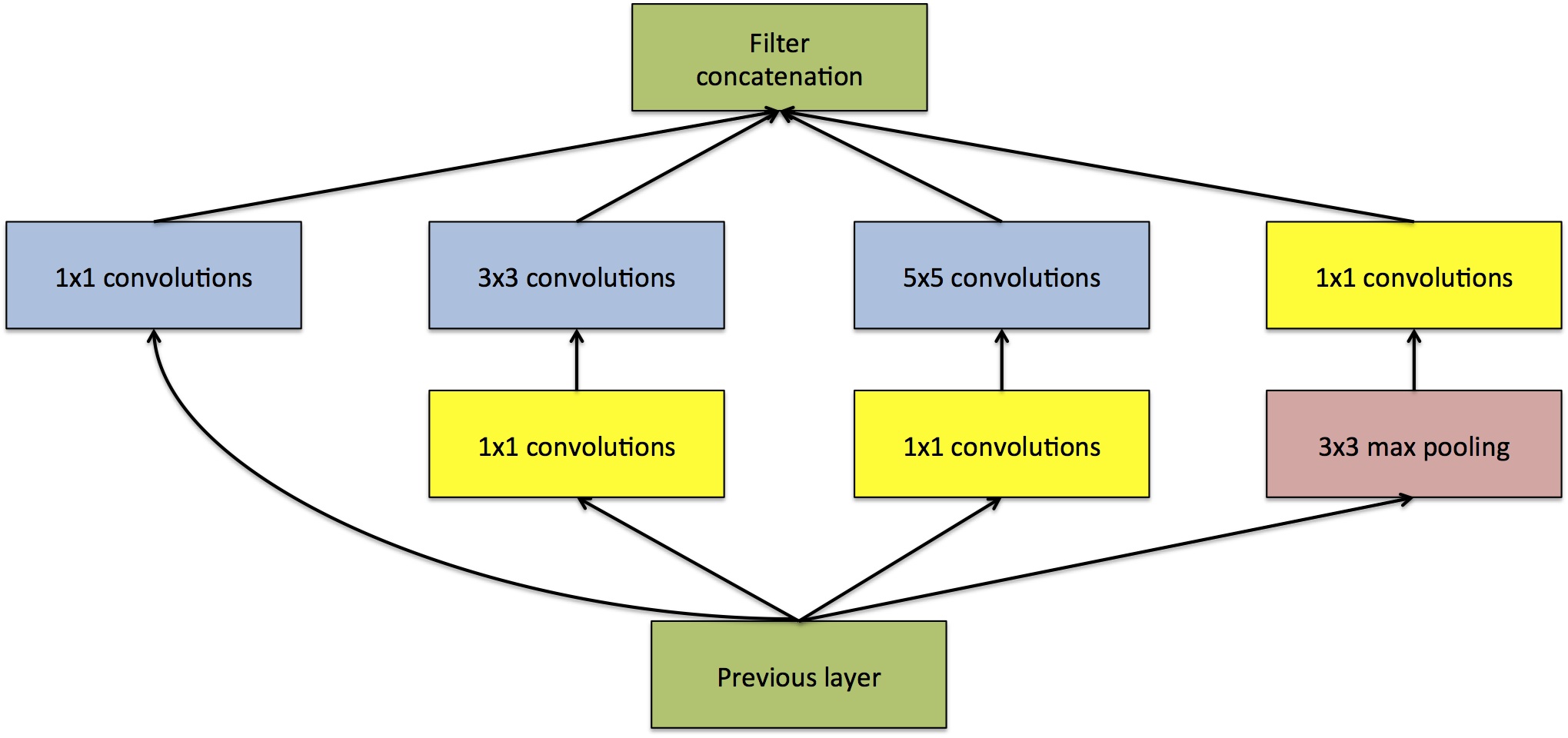} \\
(c) Original inception module~\cite{Szegedy_2015_CVPR} \\
\end{tabular}
\end{center}
\caption{The proposed deep ConvNet modules are depicted in (a) and (b), where (a) only contains multi-scale convolutional filters within each module, while (b) contains the max-pooling path, which resembles the original inception module depicted in (c) for comparison.}
\label{fig:module}
\end{figure}

In this paper, we introduce a new module for deep ConvNets composed of several multi-scale convolutional filters that are joined by a maxout activation unit, which promotes competition among these filters.  Our idea has been inspired by the recently proposed inception module~\cite{Szegedy_2015_CVPR}, which currently produces state-of-the-art results on the ILSVRC 2014 classification
and detection challenges~\cite{russakovsky2014imagenet}.  The gist of our proposal is depicted in Fig.~\ref{fig:module}, where we have the data in the input layer filtered in parallel by a set of multi-scale convolutional filters~\cite{gong2014multi,Szegedy_2015_CVPR,zagoruyko2015learning}.  Then the output of each scale of the convolutional layer passes through a batch normalisation unit (BNU)~\cite{ioffe2015batch} that weights the importance of each scale and also pre-conditions the model (note that the pre-conditioning ability of BNUs in ConvNets containing piece-wise linear activation units has recently been empirically shown in~\cite{liao2015on}).
Finally, the multi-scale filter outputs, weighted by BNU, are joined with a maxout unit~\cite{goodfellow2013maxout} that reduces the dimensionality of the joint filter outputs and promotes competition among the multi-scale filters, which prevents filter co-adaptation and allows the formation of multiple sub-networks.
We show that the introduction of our proposal module in a typical deep ConvNet produces the best results in the field for the benchmark datasets CIFAR-10~\cite{krizhevsky2009learning}, CIFAR-100~\cite{krizhevsky2009learning}, and street view house number (SVHN)~\cite{netzer2011reading}, while producing competitive results for MNIST~\cite{lecun1998gradient}.

\section{Literature Review}

\begin{figure}
\begin{center}
\begin{tabular}{c}
\includegraphics[width=0.35\textwidth]{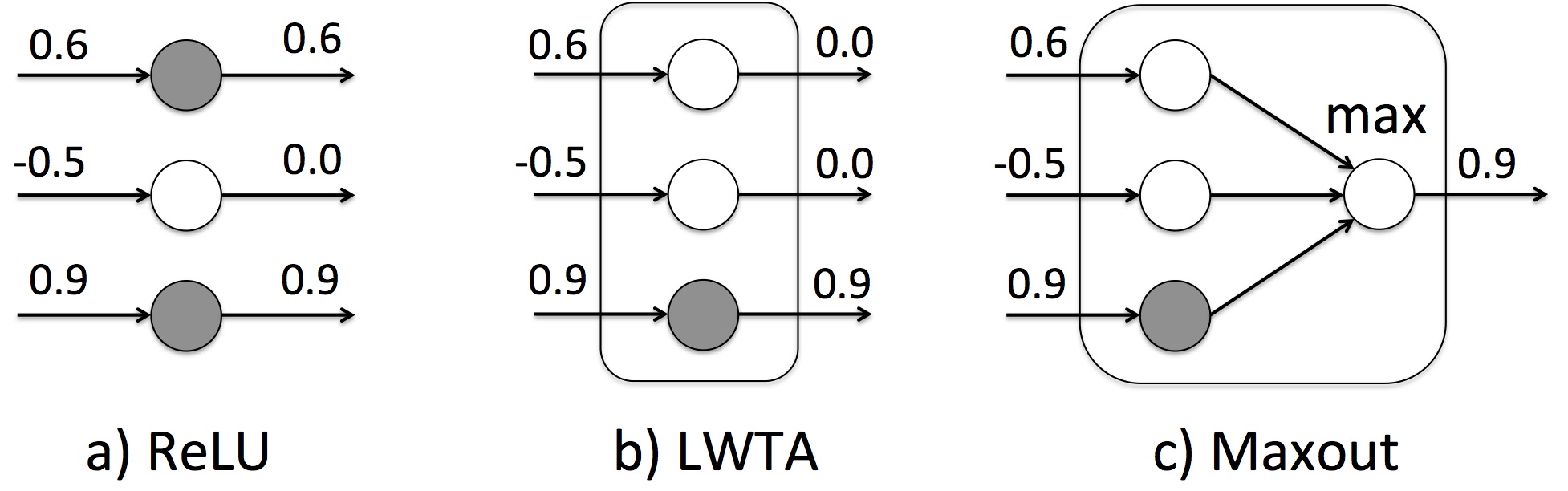}
\end{tabular}
\end{center}
\caption{Competitive activation units, where the gray nodes are the active ones, from which errors flow during backpropagation.  ReLU~\cite{glorot2011deep} (a) is active when the input is bigger than 0,  LWTA~\cite{srivastava2013compete} (b) activates only the node that has the maximum value (setting to zero the other ones), and maxout~\cite{goodfellow2013maxout} (c) has only one output containing the maximum value from the input. This figure was adapted from Fig.1 of~\cite{srivastava2014understanding}.}
\label{fig:piecewise_linear}
\end{figure}

One of the main reasons behind the outstanding performance of deep ConvNets is attributed to the use of competitive activation units in the form of piece-wise linear functions~\cite{montufar2014number,srivastava2014understanding}, such as ReLU~\cite{glorot2011deep}, maxout~\cite{goodfellow2013maxout} and LWTA~\cite{srivastava2013compete} (see Fig.~\ref{fig:piecewise_linear}).  In general, these activation functions enable the formation of sub-networks that respond consistently to similar input patterns~\cite{srivastava2014understanding},  dividing the input data points (and more generally the training space) into regions~\cite{montufar2014number}, where classifiers and regressors can be learned more effectively given that the sub-problems in each of these regions is simpler than the one involving the whole training set.  In addition, the joint training of the sub-networks present in such deep ConvNets represents a useful regularization method~\cite{glorot2011deep,goodfellow2013maxout,srivastava2013compete}.  
In practice, ReLU allows the division of the input space into two regions, but maxout and LWTA can divide the space in as many regions as the number of inputs, so for this reason, the latter two functions can estimate exponentially complex functions more effectively because of the larger number of sub-networks that are jointly trained.
An important aspect about deep ConvNets with competitive activation units is the fact that the use of batch normalization units (BNU) helps not only with respect to the convergence rate~\cite{ioffe2015batch}, but also with the pre-conditioning of the model by promoting an even distribution of the input data points, which results in the maximization of the number of the regions (and respective sub-networks) produced by the piece-wise linear activation functions~\cite{liao2015on}.  
Furthermore, training ConvNets with competitive activation units~\cite{liao2015on,srivastava2014understanding} usually involves the use of dropout~\cite{srivastava2014dropout} that consists of a regularization method that prevents filter co-adaptation~\cite{srivastava2014dropout}, which is a particularly important issue in such models, because filter co-adaptation can lead to a severe reduction in the number of the sub-networks that can be formed during training.

Another aspect of the current research on deep ConvNets is the idea of making the network deeper, which has been shown to improve classification results~\cite{goodfellow2013multi}.  However, one of the main ideas being studied in the field is how to increase the depth of a ConvNet without necessarily increasing the complexity of the model parameter space~\cite{simonyan2014very,Szegedy_2015_CVPR}.  For the Szegedy et al.'s model~\cite{Szegedy_2015_CVPR}, this is achieved with the use of $1 \times 1$ convolutional filters~\cite{lin2013network} that are placed before each local filter present in the inception module in order to reduce the input dimensionality of the filter.  In Simonyan et al.'s approach~\cite{simonyan2014very}, the idea is to use a large number of layers with convolutional filters of very small size (e.g., $3 \times 3$).  
In this work, we restrict the complexity of the deep ConvNet with the use of maxout activation units, which selects only one of the input nodes, as shown in Fig,~\ref{fig:piecewise_linear}.  

Finally, multi-scale filters in deep ConvNets is another important implementation that is increasingly being explored by several researchers~\cite{gong2014multi,Szegedy_2015_CVPR,zagoruyko2015learning}.  Essentially, multi-scale filtering follows a neuroscience model~\cite{serre2007robust} that suggests that the input image data should be processed at several scales and then pooled together, so that the deeper processing stages can become robust to scale changes~\cite{Szegedy_2015_CVPR}.  We explore this idea in our proposal, as depicted in Fig.~\ref{fig:module}, but we also argue (and show some evidence) that the multi-scale nature of the filters can prevent their co-adaptation during training.


\section{Methodology}
\label{sec:methodology}

Assume that an image is represented by $\mathbf{x}:\Omega \rightarrow \mathbb R$, where $\Omega$ denotes the image lattice, and that an image patch of size $(2k - 1) \times (2k-1)$ (for $k \in \{1,2,...,K\}$) centred at position $i \in \Omega$ is represented by $\mathbf{x}_{i \pm (k-1)}$.  The models being proposed in this paper follow the structure of the the NIN model~\cite{lin2013network}, and is in general defined as follows:
\begin{equation}
f( \mathbf{x}, \theta_f) = f_{out} \circ f_L \circ ... \circ f_2 \circ f_1(\mathbf{x}),
\label{eq:deep_conv_net}
\end{equation}
where $\circ$ denotes the composition operator, $\theta_f$ represents all the ConvNet parameters (i.e., weights and biases), $f_{out}(.)$ denotes an averaging pooling unit followed by a softmax activation function~\cite{lin2013network}, and the network has blocks represented by $l \in \{1,...,L\}$, with each block containing a composition of $N_l$ modules with $f_l( \mathbf{x}) = f_l^{(N_l)} \circ ...\circ f_l^{(2)} \circ f_l^{(1)}(\mathbf{x})$.  Each module $f_l^{(n)}(.)$ at a particular position $i \in \Omega$ of the input data for block $l$ is defined by
\begin{equation}
\begin{split}
f_l^{(n)}( \mathbf{x}_i ) =  \sigma \big (  & \gamma_{1}\mathbf{W}_{1}^{\top}\mathbf{x}_{i} + \beta_{1}, \gamma_{3}\mathbf{W}_{3}^{\top}\mathbf{x}_{i\pm 1} + \beta_{3},...,\\
& \gamma_{2k-1}\mathbf{W}_{2k-1}^{\top}\mathbf{x}_{i \pm (k-1)} + \beta_{2k-1},\\
& \gamma_p \mathbf{W}_{1}^{\top} p_{3 \times 3}( \mathbf{x}_{i \pm 1}) + \beta_p  \big ). \\
\end{split}
\label{eq:f_layers}
\end{equation}
where $\sigma(.)$ represents the maxout activation function~\cite{goodfellow2013maxout}, the convolutional filters of the module are represented by the weight matrices $\mathbf{W}_{2k-1}$ for $k \in \{1,...,K_l\}$ (i.e., filters of size $2k-1 \times 2k-1 \times \#filters$, with $\#filters$ denoting the number of $2$-D filters present in $\mathbf{W}$), which means that each module $n$ in block $l$ has $K_l$ different filter sizes and $\#filters$ different filters, $\gamma$ and $\beta$ represent the batch normalization scaling and shifting parameters~\cite{ioffe2015batch}, and $p_{3 \times 3}(\mathbf{x}_{i \pm 1})$ represents a max pooling operator on the $3 \times 3$ subset of the input data for layer $l$ centred at $i \in \Omega$, \ie $\mathbf{x}_{i \pm 1}$.

\begin{figure*}
\begin{center}
\includegraphics[width=0.8\textwidth]{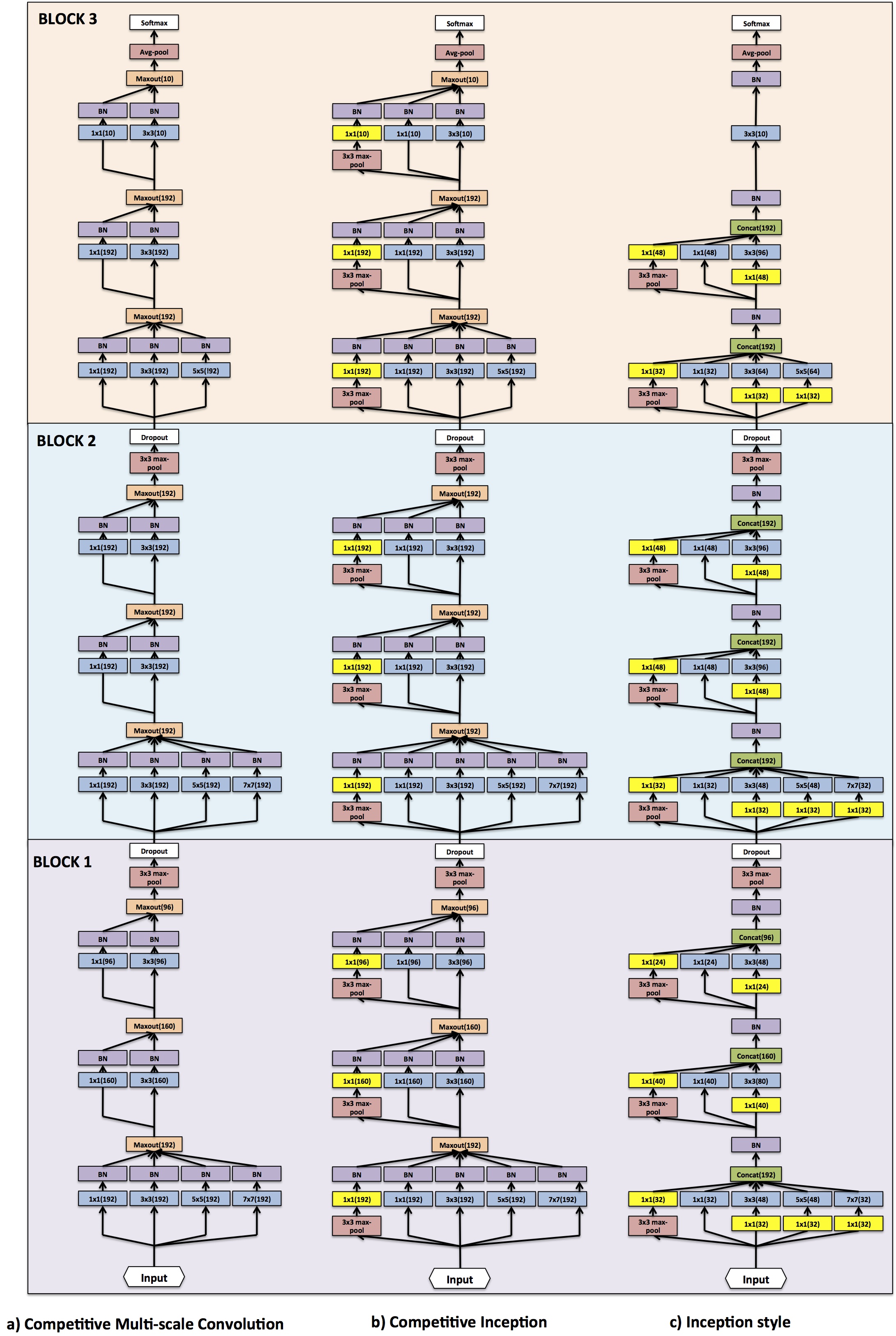} 
\end{center}
\caption{The proposed competitive multi-scale convolution (a) and competitive inception (b) networks, together with the reference inception style network (c).  In these three models, we ensure that the output of each layer has the same number of units.  Also note that: the inception style model uses ReLU~\cite{nair2010rectified} after all convolutional layers, the number of filters per convolutional node is represented by the number in brackets, and these models assume a 10-class classification problem.}
\label{fig:architecture}
\end{figure*}

Using the ConvNet module defined in (\ref{eq:f_layers}), our proposed models differ mainly in the presence or absence of the node with the max-pooling operator within the module (\ie, the node represented by $\gamma_p \mathbf{W}_{1}^{\top} p_{3 \times 3}( \mathbf{x}_{i \pm 1}) + \beta_p$).  When the module does not contain such node, 
it is called {\bf Competitive Multi-scale Convolution} (see Fig.~\ref{fig:architecture}-(a)), but when the module has the max-pooling node, then we call it {\bf Competitive Inception} (see Fig.~\ref{fig:architecture}-(b)) because of its similarity to the original inception module~\cite{Szegedy_2015_CVPR}.  The original inception module is also implemented for comparison purposes (see Fig.~\ref{fig:architecture}-(c)), and we call this model the {\bf Inception Style}, which is similar to (\ref{eq:deep_conv_net}) and (\ref{eq:f_layers}) but with the following  differences: 1) the function $\sigma(.)$ in (\ref{eq:f_layers}) denotes the concatenation of the input parameters; 2) a $1 \times 1$ convolution is applied to the input $\mathbf{x}$ before a second round of convolutions with filter sizes larger than or equal to $3 \times 3$; and 3) a ReLU activation function~\cite{glorot2011deep} is present after each convolutional layer.

An overview of all models with the structural parameters is displayed in Fig.~\ref{fig:architecture}.
Note that all models are inspired by NIN~\cite{lin2013network}, GoogLeNet~\cite{Szegedy_2015_CVPR}, and MIM~\cite{liao2015on}.  In particular, we replace the original $5 \times 5$ convolutional layers of MIM by multi-scale filters of sizes $1 \times 1$, $3 \times 3$, $5 \times 5$, and $7 \times 7$.  
For the inception style model, we ensure that the number of output units in each module is the same as for the competitive inception and competitive multi-scale convolution, and we also use a $3 \times 3$ max-pooling path in each module, as used in the original inception module~\cite{Szegedy_2015_CVPR}.
Another important point is that in general, when designing the inception style network, we follow the suggestion by Szegedy \etal~\cite{Szegedy_2015_CVPR} and include a relatively larger number of $3 \times 3$ and $5 \times 5$ filters in each module, compared to filters of other sizes (\eg, $1 \times 1$ and $7 \times 7$).  
An important distinction between the original GoogLeNet~\cite{Szegedy_2015_CVPR} and the inception style network in Fig.~\ref{fig:architecture}-(c) is the fact that we replace the fully connected layer in the last layer by a single $3 \times 3$ convolution node in the last module, followed by an average pooling and a softmax unit, similarly to the NIN model~\cite{lin2013network}.  We propose this modification to limit the number of training parameters (with the removal of the fully connected layer) and to avoid the concatenation of the nodes from different paths (\ie, maxpooling, $1 \times 1$ convolution filter, and etc.) into a number of channels that is equal to the number of classes (\ie, each channel is averaged into a single node, which is used by a single softmax unit), where the concatenation would imply that some of the paths would be directly linked to a subset of the classes.

\subsection{Competitive Multi-scale Convolution Prevent Filter Co-adaptation}
\label{sec:methodology_explanation}

The main reason being explored in the field to justify the use of competitive activation units~\cite{glorot2011deep,goodfellow2013maxout,srivastava2013compete} is the fact that they build a network formed by multiple underlying sub-networks~\cite{srivastava2014understanding}.
More clearly, given that these activation units consist of piece-wise linear functions, it has been shown that the composition of several layers containing such units, divide the input space in a number of regions that is exponentially proportional to the number of network layers~\cite{montufar2014number}, where sub-networks will be trained with the samples that fall into one of these regions, and as a result become specialised to the problem in that particular region~\cite{srivastava2014understanding}, where overfitting can be avoided because these sub-networks must share their parameters with one another~\cite{srivastava2014understanding}.
It is worth noting that these regions can only be formed if the underlying convolutional filters do not co-adapt, otherwise 
all input training samples will fall into only one region of the competitive unit, which degenerates into a simple linear transform, preventing the formation of the sub-networks.

A straightforward solution to avoid such co-adaptation can be achieved by limiting the number of training samples in a mini-batch during stochastic gradient descent.  These small batches allow the generation of ``noisy'' gradient directions during training that can activate different maxout gates, so that the different linear pieces of the activation unit can be fitted, allowing the formation of an exponentially large number of regions.  However, the drawback of this approach lies in the determination of the ``right'' number of samples per mini-batch.  A mini-batch size that is too small leads to poor convergence, and if it is too large, then it may not allow the formation of many sub-networks.  Recently, Liao and Carneiro~\cite{liao2015on} propose a solution to this problem based on the use of BNU~\cite{ioffe2015batch} that distributes the training samples evenly over the regions formed by the competitive unit, allowing the training to use different sets of training points for each region of the competitive unit, resulting in the formation of an exponential number of sub-networks.  However, there is still a potential problem with that approach~\cite{liao2015on}, which is that the underlying convolutional filters are trained using feature spaces of the same size (i.e., the underlying filters are of fixed size), which can induce the filters to co-adapt and converge to similar regions of the feature space, also preventing the formation of the sub-networks.

The competitive multi-scale convolution module proposed in this paper represents a way to fix the issue introduced above~\cite{liao2015on}.  Specifically, the different sizes of the convolutional filters within a competitive unit force the feature spaces of the filters to be different from each other, reducing the chances that these filters will converge to similar regions of the feature space.  For instance, say you have two filters of sizes $3 \times 3$ and $5 \times 5$ being joined by a competitive unit, so this means that the former filter will have a $9$-dimensional space, while the latter filter will have $16$ additional dimensions for a total of $25$ dimensions, where these new dimensions will allow the training process for the $5 \times 5$ filter to have a significantly larger feature space (\ie, for these two filters to converge to similar values, the additional $16$ dimensions will have to be pushed towards zero and the remaining $9$ dimensions to converge to the same values as the $3 \times 3$ filter).  In other words, the different filter sizes within a competitive unit imposes a soft constraint that the filters must converge to different values, avoiding the co-adaptation issue.  In some sense, this idea is similar to DropConnect~\cite{wan2013regularization}, which, during training, drops to zero the weights of randomly picked network connections with the goal of training regularization.  Nevertheless, the underlying filters will have the same size, which promotes co-adaptation even with random connections being dropped to zero.  
Compared with DropConnect that stochastically drops filter connections during training, our approach deterministically drops the border connections of a $7 \times 7$ filter (\eg, a $5 \times 5$ filter is a $7 \times 7$ filter with the $24$ border connections dropped to zero, and a $3 \times 3$ filter is a $7 \times 7$ filter with the 40 border connections forced to zero - see Fig.~\ref{fig:random_mask}).  We show in the experiments that our approach is more effective than DropConnect at the task of preventing filter co-adaptation within competitive units.

\section{Experiments}
\label{sec:experiments}

We quantitatively measure the performance of our proposed models {\bf Competitive Multi-scale Convolution} and {\bf Competitive Inception} on four computer vision/machine learning benchmark datasets: CIFAR-10\cite{krizhevsky2009learning}, CIFAR-100\cite{krizhevsky2009learning}, MNIST~\cite{lecun1998gradient} and SVHN~\cite{netzer2011reading}.
We first describe the experimental setup, then using CIFAR-10 and MNIST, we show a quantitative analysis (in terms of classification error, number of model parameters  and train/test time) of the two proposed models, the Inception Style model presented in Sec.~\ref{sec:methodology}, and two additional versions of the proposed models that justify the use of multi-scale filters, explained in Sec.~\ref{sec:methodology_explanation}.  Finally, we compare the performance of the proposed Competitive Multi-scale Convolution and Competitive Inception with respect to the current state of the art in the four benchmark datasets mentioned above.

The CIFAR-10~\cite{krizhevsky2009learning} dataset contains 60000 images of 10 commonly seen object categories (e.g., animals, vehicles, etc.), where 50000 images are used for training and the rest 10000 for testing, and all 10 categories have equal volume of training and test images. 
The images of CIFAR-10 consist of $32 \times 32$-pixel RGB images, where the objects are well-centered in the middle of the image.
The CIFAR-100~\cite{krizhevsky2009learning} dataset extends CIFAR-10 by increasing the number of categories to 100, whereas the total number of images remains the same, so 
the CIFAR-100 dataset is considered as a harder classification problem than CIFAR-10 since it contains 10 times less images per class and 10 times more categories.
The well-known MNIST~\cite{lecun1998gradient} dataset contains $28 \times 28$ grayscale images comprising 10 handwritten digits (from $0$ to $9$), where the dataset is divided into 60000 images for training and 10000 for testing, but note that the number of images per digit is not uniformly distributed.
Finally, the Street View House Number (SVHN)~\cite{netzer2011reading} is also a digit classification benchmark dataset that contains 600000 $32 \times 32$ RGB images of printed digits (from $0$ to $9$) cropped from pictures of house number plates.
The cropped images is centered in the digit of interest, but nearby digits and other distractors are kept in the image. 
SVHN has three sets: training, testing sets and a extra set with 530000 images that are less difficult and can be used for helping with the training process.
We do not use data augmentation in any of the experiments, and we only compare our results with other methods that do not use data augmentation.

In all these benchmark datasets we minimize the softmax loss function present in the last layer of each model for the respective classification in each dataset, and we report the results as the proportion of misclassified test images, which is the standard way of comparing algorithms in these benchmark datasets.
The reported results are generated with the models trained using an initial learning rate of 0.1 and following a multi-step decay to a final learning rate of 0.001 (in 80 epochs for CIFAR-10 and CIFAR-100, 50 epochs for MNIST, and 40 epochs for SVHN).  The stopping criterion is determined by the convergence observed in the error on the validation set. The mini-batch size for CIFAR-10, CIFAR-100, and MNIST datasets is 100, and 128 for SVHN dataset. The momentum and weight decay are set to standard values 0.9 and 0.0005, respectively. 
For each result reported, we compute the mean and standard deviation of the test error from five separately trained models, where for each model, we use the same training set and parameters (e.g., the learning rate sequence, momentum, etc.), and we change only the random initialization of the filter weights and randomly shuffle the training samples.

We use the GPU-accelerated ConvNet library MatConvNet~\cite{vedaldi15matconvnet} to perform the experiments specified in this paper.
Our experimental environment is a desktop PC equipped with i7-4770 CPU, 24G memory and a 12G GTX TITAN X graphic card.
Using this machine, we report the mean training and testing times of our models.

\subsection{Model Design Choices}
\label{sec:model_design_choices}

In this section, we show the results from several experiments that show the design choices for our models, where we provide comparisons in terms of their test errors, the number of parameters involved in the training process and the training and testing times.  Tables~\ref{tab:cifar10_exp_component}~and~\ref{tab:mnist_exp_component} show the results on CIFAR-10 and MNIST for the models {\bf Competitive Multi-scale Convolution}, {\bf Competitive Inception}, and {\bf Inception Style} models, in addition to other models explained below.  Note that all models in Tables~\ref{tab:cifar10_exp_component}~and~\ref{tab:mnist_exp_component} are constrained to have the same numbers of input channels and output channels in each module, and all networks contain three blocks~\cite{lin2013network}, each with three modules (so there is a total of nine modules in each network), as shown in Fig.~\ref{fig:architecture}.

\begin{figure}
\begin{center}
\includegraphics[width=0.7\columnwidth]{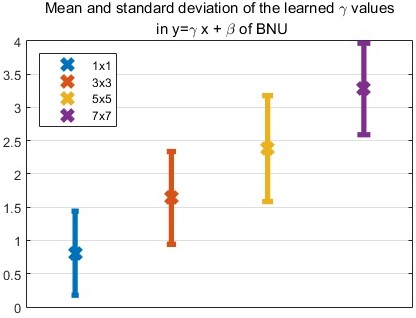} 
\end{center}
\caption{Mean and standard deviation of the learned $\gamma$ values in the batch normalisation unit of (\ref{eq:f_layers}) for the Competitive Multi-scale Convolution model on CIFAR-10.  This result provides an estimate of the importance placed on each filter by the training procedure.}
\label{fig:filter_importance_measurement_gamma}
\end{figure}

We argue that the multi-scale nature of the filters within the competitive module is important to avoid the co-adaptation issue explained in Sec.~\ref{sec:methodology_explanation}.  We assess this importance by comparing both the number of parameters and the test error results between the proposed models and the model {\bf Competitive Single-scale Convolution}, which has basically the same architecture as the Competitive Multi-scale Convolution model represented in Fig.~\ref{fig:architecture}-(a), but with the following changes: the first two blocks contain four sets of $7 \times 7$ filters in the first module, and in the second and third modules, two sets of $3 \times 3$ filters; and the third block has three filters of size $5 \times 5$ in the first module, followed by two modules with two $3 \times 3$ filters.  Notice that this configuration implies that we replace the multi-scale filters by the filter of the largest size of the module in each node, which is a configuration similar to the recently proposed MIM model~\cite{liao2015on}.  The configuration for the Competitive Single-scale Convolution has around two times more parameters than the Competitive Multi-scale Convolution model and takes longer to train, as displayed in Tables~\ref{tab:cifar10_exp_component}~and~\ref{tab:mnist_exp_component}.   The idea behind the use of the largest size filters within each module is based on the results obtained from the training of the batch normalisation units of the Competitive Multi-scale Convolution modules, which indicates that the highest weights (represented by $\gamma$ in (\ref{eq:f_layers})) are placed in the largest size filters within each module, as shown in Fig.~\ref{fig:filter_importance_measurement_gamma}.  The classification results of the Competitive Single-scale Convolution, shown in Tables~\ref{tab:cifar10_exp_component}~and~\ref{tab:mnist_exp_component}, demonstrate that it is consistently inferior to the Competitive Multi-scale Convolution model.

\begin{figure}
\begin{center}
\includegraphics[width=0.7\columnwidth]{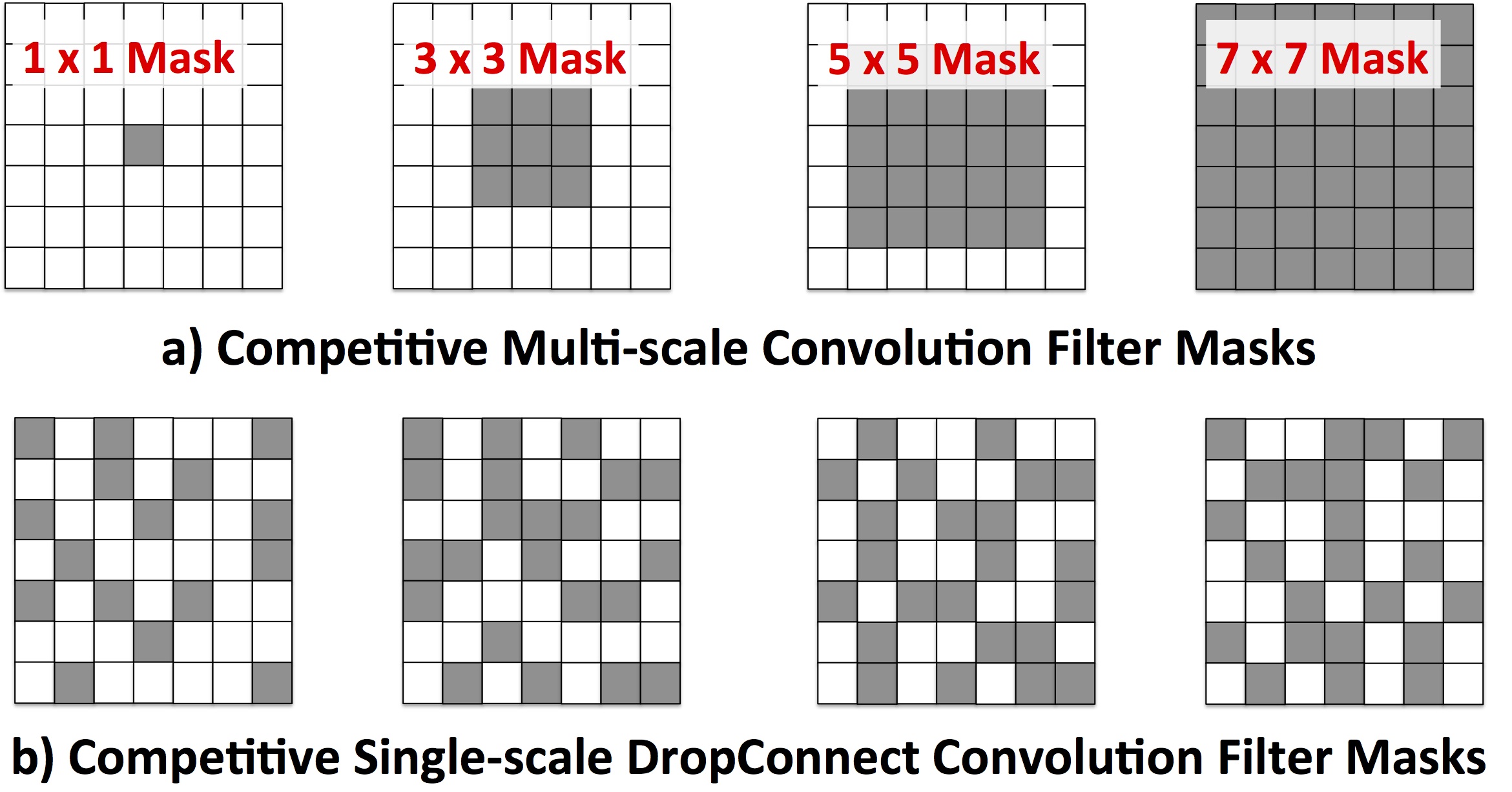} 
\end{center}
\caption{The Competitive Multi-scale Convolution module has filters of size $1 \times 1$, $3 \times 3$, $5 \times 5$, and $7 \times 7$, which is equivalent to having four $7 \times 7$ filters (with a total of 196 weights) with the masks in (a), where the number of deterministically masked out (or dropped) weights is 112.  Using a DropConnect rate of  $112/196 \approx 0.57$, a possible set of randomly dropped weights is shown in (b).  Note that even though the proportion and number of weights dropped in (a) and (b) are the same, the deterministic or stochastic masking of the weights make a difference in the performance, as explained in the paper.}
\label{fig:random_mask}
\end{figure}

Another important point that we test in this section is the relevance of dropping connections in a deterministic or stochastic manner when training the competitive convolution modules.  
Recall that the one of the questions posed in Sec.~\ref{sec:methodology_explanation} is if the deterministic masking provided by our proposed Competitive Multi-scale Convolution module is more effective at avoiding filter co-adaptation than the stochastic masking provided by DropConnect~\cite{wan2013regularization}.
We run a quantitative analysis of the {\bf Competitive DropConnect Single-scale Convolution}, where we take the Competitive Single-scale Convolution proposed before and randomly drop connections using a rate, which is computed such that it has on average the same number of parameters to learn in each round of training as the Competitive Multi-scale Convolution, but notice that the Competitive DropConnect Single-scale Convolution has in fact the same number of parameters as the Competitive Single-scale Convolution.  Using Fig.~\ref{fig:random_mask}, we see that the DropConnect rate is 0.57 for the module 1 of blocks 1 and 2 specified in Fig.~\ref{fig:architecture}.
The results in Tables~\ref{tab:cifar10_exp_component}~and~\ref{tab:mnist_exp_component} show that it has around two times more parameters, takes longer to train and performs significantly worse than the Competitive Multi-scale Convolution model.

Finally, the reported training and testing times in Tables~\ref{tab:cifar10_exp_component}~and~\ref{tab:mnist_exp_component} show a clear relation between the number of model parameters and those times.  

\begin{table}[]
\begin{center}
\resizebox{1.0\columnwidth}{!}{
\begin{tabular}[]{lrrrr}
\hline
Method & No. of Params & Test Error & Train Time & Test Time\\ 
& & (mean $\pm$ std dev) & (h) & (ms)\\
\hline
Competitive Multi-scale  & $4.48 \text{ M}$ & $6.87 \pm 0.05 \%$  & 6.4 h & 2.7 ms \\
Convolution &&&& \\
Competitive Inception & $4.69\text{ M}$ & $7.13 \pm 0.31 \%$  & 7.6 h & 3.1 ms \\
&&&&\\
Inception Style & $0.61 \text{ M}$ & $8.50 \pm 0.06 \%$  & 3.9 h & 1.5 ms \\
&&&&\\
Competitive Single-scale & $9.35 \text{ M}$ & $7.15 \pm 0.12 \%$  & 8.0 h & 3.2 ms \\
Convolution &&&& \\
Competitive DropConnect  & $9.35 \text{ M}$ & $9.12 \pm 0.17 \%$  & 7.7 h & 3.1 ms \\
 Single-scale Convolution &&&& \\
\hline
\end{tabular}
}
\end{center}
\caption{Results on CIFAR-10 of the proposed models, in addition to the Competitive Single-scale Convolution and Competitive DropConnect Single-scale Convolution that test our research questions posed in Sec.~\ref{sec:methodology_explanation}.}
\label{tab:cifar10_exp_component}

\end{table}

\begin{table}[]
\begin{center}
\resizebox{1.0\columnwidth}{!}{
\begin{tabular}[]{lrrrr}
\hline
Method & No. of Params & Test Error & Train Time & Test Time\\ 
& & (mean $\pm$ std dev) & (h) & (ms)\\
\hline
Competitive Multi-scale & $1.13 \text{ M}$ & $0.33 \pm 0.04\%$ & 1.5 h & 0.8 ms \\
Convolution &&&& \\
Competitive Inception & $1.19 \text{ M}$ & $0.40 \pm 0.02\% $ & 1.9 h & 1.0 ms \\
&&&& \\
Inception Style & $ 0.18 \text{ M}$ & $ 0.44  \pm  0.01   \%$ & 1.4 h & 0.7 ms \\
&&&& \\
Competitive Single-scale  & $2.39 \text{ M}$ & $0.37 \pm 0.03\% $ & 1.7 h & 0.9 ms  \\
Convolution &&&& \\
Competitive DropConnect  &  $2.39 \text{ M}$ & $  0.35 \pm 0.03 \%$ & 1.6 h & 0.9 ms \\
Single-scale Convolution &&&& \\
\hline
\end{tabular}
}
\end{center}
\caption{Results on MNIST of the proposed models, in addition to the Competitive Single-scale Convolution and Competitive DropConnect Single-scale Convolution that test our research questions posed in Sec.~\ref{sec:methodology_explanation}.}
\label{tab:mnist_exp_component}
\end{table}

\subsection{Comparison with the State of the Art}
\label{sec:comparison_state_of_the_art}

We now show the performances of the proposed Competitive Multi-scale and Competitive Inception Convolution models on CIFAR-10, CIFAR-100, MNIST and SVHN, and compare them with the current state of the art in the field, which can be listed as follows.  {\bf Stochastic Pooling}~\cite{zeiler2013stochastic} proposes a regularization based on a replacement of the deterministic pooling (e.g., max or average pooling) by a stochastic procedure, which randomly selects the activation within each pooling region according to a multinomial distribution, estimated from the activation of the pooling unit. {\bf Maxout Networks}~\cite{goodfellow2013maxout} introduces a piece-wise linear activation unit that is used together with dropout training~\cite{srivastava2014dropout} and is introduced in Fig.~\ref{fig:piecewise_linear}-(c).
The {\bf Network in Network} (NIN)~\cite{lin2013network} model consists of the introduction of multilayer perceptrons as activation functions to be placed between convolution layers, and the replacement of a final fully connected layer by average pooling, where the number of output channels represent the final number of classes in the classification problem.  {\bf Deeply-supervised nets}~\cite{lee2014deeply} introduce explicit training objectives to all hidden layers, in addition to the back-propagated errors from the last softmax layer.
The use of a recurrent structure that replaces the purely feed-forward structure in ConvNets is explored by the model {\bf RCNN}~\cite{liang2015recurrent}.  An extension of the NIN model based on the use of maxout activation function instead of the multilayer perceptron is introduced in the {\bf MIM} model~\cite{liao2015on}, which also shows that the use of batch normalization units are crucial for allowing an effective training of several single-scale filters that are joined by maxout units.  Finally, 
the {\bf Tree based Priors}~\cite{srivastava2013discriminative} model proposes a training method for classes with few samples, using a generative prior that is learned from the data and shared between related classes during the model learning.

The comparison on CIFAR-10~\cite{krizhevsky2009learning} dataset is shown in Tab.~\ref{tab:cifar10_exp}, where results are sorted based on the performance of each method, and the results of our proposed methods are highlighted.
The results on CIFAR-100\cite{krizhevsky2009learning} dataset are displayed in Tab.\ref{tab:cifar100_exp}.
Table~\ref{tab:mnist_exp} shows the results on MNIST~\cite{lecun1998gradient}, where it is worth reporting that the best result (over the five trained models) produced by our Competitive Multi-scale Convolution model is a test error of $0.29\%$, which is better than the single result from Liang and Hu~\cite{liang2015recurrent}.
Finally, the comparison on SVHN\cite{netzer2011reading} dataset is shown in Table~\ref{tab:svhn_exp}, where two out of the five models show test error results of $1.69\%$.

\begin{table}[]
\begin{center}
\resizebox{1.0\columnwidth}{!}{
\begin{tabular}[]{lr}
\hline
Method &  Test Error (mean $\pm$ standard deviation)\\
\hline
{\bf Competitive Multi-scale Convolution} & $6.87 \pm 0.05 \%$ \\
{\bf Competitive Inception} & $7.13 \pm 0.31 \%$ \\
MIM~\cite{liao2015on} & $8.52 \pm 0.20 \%$ \\
RCNN-160~\cite{liang2015recurrent} & $8.69\%$ \\
Deeply-supervised nets~\cite{lee2014deeply} & $9.69\%$ \\
Network in Network~\cite{lin2013network} & $10.41\%$ \\
Maxout Networks~\cite{goodfellow2013maxout} & $11.68\%$ \\
Stochastic Pooling~\cite{zeiler2013stochastic} &  $15.13\%$ \\
\hline
\end{tabular}
}
\end{center}
\caption{Comparison in terms of classification error between our proposed models (highlighted) and the state-of-the-art methods on CIFAR-10~\cite{krizhevsky2009learning}. }
\label{tab:cifar10_exp}
\end{table}

\begin{table}[]
\begin{center}
\resizebox{1.0\columnwidth}{!}{
\begin{tabular}[]{lr}
\hline
Method &  Test Error (mean $\pm$ standard deviation)\\
\hline
{\bf Competitive Multi-scale Convolution} & $27.56 \pm 0.49 \%$ \\
{\bf Competitive Inception} & $28.17 \pm 0.25 \%$ \\
MIM~\cite{liao2015on}  & $29.20 \pm 0.20 \%$ \\
RCNN-160~\cite{liang2015recurrent} & $31.75\%$ \\
Deeply-supervised nets~\cite{lee2014deeply} & $34.57\%$ \\
Network in Network~\cite{lin2013network} & $35.68\%$ \\
Tree based Priors~\cite{srivastava2013discriminative} & $36.85\%$ \\
Maxout Networks~\cite{goodfellow2013maxout} & $38.57\%$ \\
Stochastic Pooling~\cite{zeiler2013stochastic} &  $42.51\%$ \\
\hline
\end{tabular}
}
\end{center}
\caption{Comparison in terms of classification error between our proposed models (highlighted) and the state-of-the-art methods on CIFAR-100~\cite{krizhevsky2009learning}. }
\label{tab:cifar100_exp}
\end{table}

\begin{table}[]
\begin{center}
\resizebox{1.0\columnwidth}{!}{
\begin{tabular}[]{lr}
\hline
Method &  Test Error (mean $\pm$ standard deviation)\\
\hline
RCNN-96~\cite{liang2015recurrent} & $0.31\%$ \\
{\bf Competitive Multi-scale Convolution} & $0.33 \pm 0.04\%$ \\
MIM~\cite{liao2015on}  & $0.35 \pm 0.03\%$ \\ 
Deeply-supervised nets~\cite{lee2014deeply} & $0.39\%$ \\
{\bf Competitive Inception} & $0.40 \pm 0.02\% $ \\
Network in Network~\cite{lin2013network} & $0.45\%$ \\
Conv. Maxout+Dropout~\cite{goodfellow2013maxout} & $0.47\%$ \\
Stochastic Pooling~\cite{zeiler2013stochastic} & $0.47\%$ \\
\hline
\end{tabular}
}
\end{center}
\caption{Comparison in terms of classification error between our proposed models (highlighted) and the state-of-the-art methods on MNIST~\cite{lecun1998gradient}.}
\label{tab:mnist_exp}
\end{table}

\begin{table}[]
\begin{center}
\resizebox{\columnwidth}{!}{
\begin{tabular}[]{lr}
\hline
Method &  Test Error (mean $\pm$ standard deviation)\\
\hline
{\bf Competitive Multi-scale Convolution} & $1.76 \pm 0.07\%$ \\
RCNN-192~\cite{liang2015recurrent} & $1.77\%$ \\
{\bf Competitive Inception Convolution} & $ 1.82 \pm 0.05\%$ \\
Deeply-supervised nets~\cite{lee2014deeply} & $1.92\%$ \\
Drop-connect~\cite{wan2013regularization} & $1.94\%$  \\
MIM~\cite{liao2015on} & $1.97 \pm 0.08\%$ \\ 
Network in Network~\cite{lin2013network} & $2.35\%$ \\
Conv. Maxout+Dropout~\cite{goodfellow2013maxout} & $2.47\%$ \\
Stochastic Pooling~\cite{zeiler2013stochastic} & $2.80\%$ \\
\hline
\end{tabular}
}
\end{center}
\caption{Comparison in terms of classification error between our proposed models (highlighted) and the state-of-the-art methods on SVHN~\cite{netzer2011reading}.}
\label{tab:svhn_exp}
\end{table}

\section{Discussion and Conclusions}
\label{sec:conclusions}

In terms of the model design choices in Sec.~\ref{sec:model_design_choices}, we can see that the proposed Competitive Multi-scale Convolution produces more accurate classification results than the proposed Competitive Inception.  Given that the main difference between these two models is the presence of the max-pooling path within each module, we can conclude that this path does not help with the classification accuracy of the model.
The better performance of both models with respect to the Inception Style model can be attributed to the maxout unit that induces competition among the underlying filters, which helps more the classification results when compared with the collaborative nature of the Inception module.
Considering model complexity, it is important to notice that the relation between the number of parameters and training and testing times is not linear, where even though the Inception Style model has 10$\times$ fewer parameters, it trains and tests 2 to 1.5$\times$ faster than the proposed Competitive Multi-scale Convolution and Competitive Inception models.

When answering the questions posed in Sec.~\ref{sec:methodology_explanation}, we assume that classification accuracy is a proxy for measuring the co-adaptation between filters within a single module, where the intuition is that if the filters joined by a maxout activation unit co-adapt and become similar to each other, a relatively small number of large regions in the input space will be formed, which results in few sub-networks to train, with each sub-network becoming less specialized to its region~\cite{montufar2014number,srivastava2014understanding}.  We argue that the main consequence of that is a potential lower classification accuracy, depending on the complexity of the original classification problem.
Using this assumption, we note from Tables~\ref{tab:cifar10_exp_component}~and~\ref{tab:mnist_exp_component} that  
the use of multi-scale filters within a competitive module is in fact important to avoid the co-adaptation of the filters, as shown by the more accurate classification results of the Multi-scale, compared to the Single-scale model. 
Furthermore, the use of deterministic, as opposed to stochastic, mapping also appears to be more effective in avoiding filter co-adaptation given the more accurate classification results of the former mapping.  Nevertheless, the reason behind the worse performance of the stochastic mapping may be due to the fact that DropConnect has been designed for the fully connected layers only~\cite{wan2013regularization}, while our test bed for the comparison is set in the convolutional filters. To be more specific, we think that a fully connected layer usually encapsulates hundreds to thousands of weights for inputs of similar scale of dimensions, thus a random dropping on a subset of weight elements can hardly change the distribution of the outputs pattern. However, the convolution filters are of small dimensions, and each of our maxout unit controls 4 to 5 filters at most, so such masking scheme over small weights matrix could result in ``catastrophic forgetting''~\cite{mccloskey1989catastrophic} which explains why the Competitive DropConnect Single-scale Convolution performs even worse than Competitive Single-scale Convolution on CIFAR-10.

We also run an experiment that assesses whether filters of larger size within a competitive module can improve the classification accuracy at the expense of having a larger number of parameters to train.  We test the inclusion of two more filters of sizes $9 \times 9$ and $11 \times 11$ in module 1 of blocks 1 and 2, and two more filter sizes $7 \times 7$  and $9 \times 9$ in module 1 of block 3 (see Fig.~\ref{fig:architecture}).  The classification result obtained is $7.36 \pm 0.16 \%$ on CIFAR-10, and number of model parameters is 13.11 M.  This experiment shows that increasing the number of filters of larger sizes do not necessarily help improve the classification results.
An important modification that can be suggested for our proposed Competitive Multi-scale Convolution model is the replacement of the maxout by ReLU activation, where only the largest size filter of each module is kept and all other filters are removed.  One can argue that such model is perhaps less complex (in terms of the number of parameters) and probably as accurate as the proposed model.  However, the results we obtained with such model on CIFAR-10 show that this model has  3.28 M parameters (\ie, just slightly less complex than the proposed models, as shown in Tab.~\ref{tab:cifar10_exp_component}) and has a classification test error of $8.16 \pm 0.15\%$, which is significantly larger than for our proposed models.  On MNIST, this model has 0.81 M parameters and produces a classification error of $0.37 \pm 0.05 \%$, which also shows no advantage over the proposed models.

The comparisons with the state of the art in Tables~\ref{tab:cifar10_exp}-~\ref{tab:svhn_exp} of Sec.~\ref{sec:comparison_state_of_the_art} show that the proposed Competitive Multi-scale Convolution model produces the best results in the field for three out of the four considered datasets.  However, note that this comparison is not strictly fair to us because we run a five-model validation experiment (using different model initializations and different sets of mini batches for the stochastic gradient descent), which provides a more robust performance assessment of our method.  In contrast, most of the methods in the field only show one single result of their performance.  If we consider only the best result out of the five results in the experiment, then our Competitive Multi-scale Convolution model has the best results in all four datasets (with, for example, $0.29\%$ on MNIST and $1.69\%$ on SVHN).  An analysis of these results also allows us to conclude that the main competitors of our approach are the MIM~\cite{liao2015on} and RCNN~\cite{liang2015recurrent} models, where the MIM method is quite related to our approach, but the RCNN method follows a quite different strategy.

In this paper, we show the effectiveness of using competitive units on modules that contain multi-scale filters.  
We argue that the main reason of the superior classification results of our proposal, compared with the current state of the art in several benchmark datasets, lies in the following points: 1) the deterministic masking implicitly used by the multi-scale filters avoids the issue of filter co-adaptation;  2) the competitive unit that joins the underlying filters and the batch normalization units promote the formation of a large number of sub-networks that are specialized in the classification problem restricted to a small area of the input space and that are regularized by the fact that they are trained together within the same model; and 3) the maxout unit allows the reduction of the number of parameters in the model.
It is important to note that such modules can be applied in several types of deep learning networks, and we plan to apply it to other types of models, such as the recurrent neural network~\cite{liang2015recurrent}.

{\small
\bibliographystyle{ieee}
\bibliography{egbib,egbib2}
}

\end{document}